\documentclass[letterpaper, 10 pt, conference]{ieeeconf}  

\IEEEoverridecommandlockouts                              

\title{\LARGE \bf
    TOAST: Stochastic Robot Action Tokenization  \\ 
    for Autoregressive Vision-Language-Action Models
}

\author{
Keisuke Shirai, Tomohiro Motoda, Hanbit Oh, Ryoichi Nakajo, \\
Roman Mykhailyshyn, Ryo Hanai, Shotaro Miwa, Yukiyasu Domae \\
AIST
\thanks{Code and videos are avaiable at \url{https://kskshr.github.io/toast}.}
}

\usepackage{graphicx}
\usepackage{amsmath}
\usepackage{amssymb}
\usepackage{color}
\usepackage{xspace}
\usepackage[normalem]{ulem}
\usepackage{subcaption}
\usepackage{multirow}
\usepackage{url}
\usepackage{booktabs}
\usepackage{hyperref}
\usepackage{cite}

\newcommand{\proposed}{TOAST\xspace}

\newcommand{\fast}{FAST\xspace}

\newcommand{\bfuline}[1]{\uline{\textbf{#1}}}

\newcommand{\secref}[1]{Section~\ref{#1}}

\newcommand{\tabref}[1]{{Table~\ref{#1}}}
\newcommand{\figref}[1]{{Figure~\ref{#1}}}

\begin{document}

\maketitle
\thispagestyle{empty}
\pagestyle{empty}


\begin{abstract}
Autoregressive Vision-Language-Action models often represent continuous robot actions as discrete token sequences, enabling action prediction with standard next-token objectives. FAST has substantially improved this representation by compactly encoding action  containing diverse temporal frequencies into relatively few tokens. However, while such compression reduces the number of action tokens required for autoregressive prediction, it does not necessarily improve the efficiency of policy learning from limited demonstrations. In particular, FAST typically assigns a single deterministic tokenization to each quantized action sequence, although multiple token sequences can represent and decode to the same robot motion. We investigate whether exploiting this representational redundancy can improve policy learning. In this paper, we propose \uline{TO}kenization of \uline{A}ction Sequences with \uline{ST}ochastic Sampling (\proposed), a stochastic action tokenization method that samples alternative tokenizations of the same quantized action sequence during policy training. This diversifies the discrete supervision while preserving the underlying robot action and requires no additional demonstrations. Experiments on LIBERO show that \proposed consistently improves over its deterministic counterpart, with the improvement increasing as training data decreases, achieving a 6.8 point gain in success rate when only $1/16$ of training data is available. Across four real-robot manipulation tasks, \proposed further improves mean success rate by $15.8$ points over the deterministic counterpart. These results demonstrate the effectiveness of stochastic action tokenization for autoregressive robot policy learning, particularly when training data are limited.
\end{abstract}

\section{Introduction}
\label{sec:introduction}
Vision-language-action models (VLAs), backed by the success of imitation learning~\cite{osa2018algorithmic}, have been drawing attention as an approach for realizing general-purpose robot policies~\cite{zitkovich2023rt2,reed2022gato,ghosh2024octo,kim2024openvla,black2025pi0,nvidia2025gr00tn1,liu2025rdt1b,lee2025molmoact}. Autoregressive VLAs~\cite{zitkovich2023rt2,kim2024openvla,pertsch2025fast,lee2025molmoact} are policies that use a backbone multimodal large language model (MLLM) to predict action sequences as tokens via next-token prediction. Autoregressive VLAs offer advantages over flow matching-based ones~\cite{black2025pi0,nvidia2025gr00tn1} of having a simpler architecture~\cite{liang2025discrete}, leveraging MLLM's pre-trained knowledge effectively~\cite{liu2026hybridvla}, and realizing faster convergence~\cite{pertsch2025fast}. In building autoregressive VLAs, the tokenizer determines what the policy has to predict, and recent work has explored efficient compression of action chunks in two ways: learned vector quantization, as in VQ-VLA~\cite{wang2025vqvla}, or training-free frequency-based compression, as in FAST~\cite{pertsch2025fast}, which applies byte pair encoding (BPE)~\cite{sennrich2016neural} to the quantized coefficients of a discrete cosine transform. Both make prediction easier by shortening the token sequence.

\begin{figure}[t]
    \centering
    \includegraphics[width=0.95\linewidth]{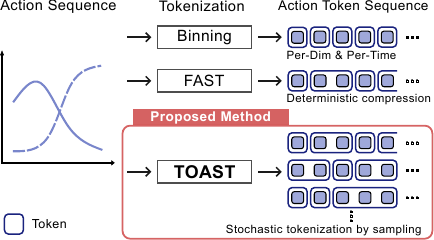}
    \caption{We propose \proposed, a novel framework that introduces stochastic action tokenization to enable efficient policy learning with supervision diversification while preserving the underlying action.}
    \label{fig:overview}
\end{figure}

However, the tokenization boundaries produced by frequency-based compression such as BPE are not necessarily physically meaningful for robot actions. Once the vocabulary contains tokens spanning multiple quantized action values, the same quantized action sequence can be represented by different token sequences, all of which decode to exactly the same motion. Deterministic tokenizers nevertheless select a single tokenization according to corpus-level frequency statistics, and the policy is trained under cross-entropy to reproduce only that sequence. A similar ambiguity has long been recognized in natural language processing, where stochastic tokenization~\cite{kudo2018subword,provilko2020bpedropout,wang2021multiview} mitigates over-reliance on a single segmentation and has shown particular benefits when training data are limited. This issue can be especially important in robot learning, where demonstrations are costly to collect. When only limited training data are available, repeatedly supervising each action sequence with the same deterministic tokenization may encourage the policy to overfit to specific token patterns or segmentation boundaries that are not uniquely tied to the underlying motion.

To address this issue, we propose \bfuline{TO}kenization of \bfuline{A}ction Sequences with \bfuline{ST}ochastic Sampling (\proposed), the first stochastic action tokenization framework for autoregressive robot policy learning. As shown in \figref{fig:overview}, \proposed samples multiple tokenizations of the same action sequence during training, exposing the policy to different discrete representations while preserving the underlying motion. Moreover, because multiple valid token sequences can be obtained from each demonstration without collecting additional robot trajectories, \proposed increases the diversity of supervision available from limited data. We therefore hypothesize that stochastic action tokenization becomes increasingly beneficial when fewer demonstrations are available. We instantiate this framework with a unigram tokenizer that enables probabilistic segmentation, together with dimension-major flattening for frequency-based action sequence compression.

We evaluate this hypothesis through both controlled simulation and real-world manipulation task experiments. The results show that stochastic tokenization consistently improves policy performance over its deterministic counterpart across training-set sizes, flattening orders, and vocabulary corpora. Notably, the relative benefit of stochastic tokenization is greater when fewer demonstrations are available, indicating improved data efficiency in policy learning. We further observe substantial gains on real-robot manipulation tasks, where collecting demonstrations is particularly expensive. These results show that stochastic action tokenization provides a simple and effective means of extracting more useful supervision from limited robot demonstrations.

Our contributions are summarized as follows:
\begin{itemize}
\item We identify deterministic action tokenization as a source of over-reliance on a single discrete representation and formulate stochastic tokenization as a representation-level regularization effect for autoregressive VLAs.
\item We introduce \proposed, which samples equivalent tokenizations of the same motion during training, yielding more diverse supervision from each demonstration without additional robot data.
\end{itemize}

\section{Related work}
\label{sec:related_work}
\textbf{Vision-language-action models and action tokenization.}
Inspired by advances in foundation models for natural language processing and computer vision~\cite{hurst2024gpt,comanici2025gemini,beyer2024paligemma,bai2025qwen3,deitke2025molmo}, Vision-Language-Action (VLA) models have emerged as a promising approach to generalist robot policy learning~\cite{reed2022gato,brohan2023rt1,zitkovich2023rt2,ghosh2024octo, kim2024openvla,black2025pi05,nvidia2025gr00tn1,geminiroboticsteam2025geminiroboticsbringingai}. Autoregressive VLAs formulate action prediction as next-token prediction~\cite{zitkovich2023rt2,kim2024openvla,pertsch2025fast,lee2025molmoact}, making the representation of continuous robot actions as discrete tokens an important design choice. Early approaches use uniform binning to discretize each action dimension and timestep independently~\cite{zitkovich2023rt2,kim2024openvla}. More recent methods exploit temporal or cross-dimensional structure to obtain more compact action representations. BEAST~\cite{zhou2025beast} represents action sequences using B-spline control points, while vector-quantization-based methods~\cite{lee2024behavior,liu2026faster}, such as VQ-VLA~\cite{wang2025vqvla}, learn discrete latent representations of continuous actions. FAST~\cite{pertsch2025fast} instead provides a training-free tokenizer by applying the discrete cosine transform (DCT) followed by byte pair encoding (BPE)~\cite{sennrich2016neural}. Our work builds on this frequency-based approach, but focuses on stochastic tokenization of action sequences rather than introducing another compression scheme.

\begin{figure*}[t]
    \centering
    \includegraphics[width=1.0\linewidth]{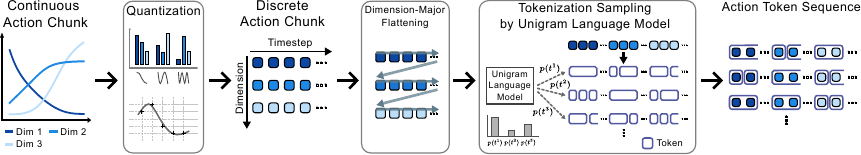}
    \caption{\textbf{\proposed tokenization pipeline.} Given a discrete action chunk after quantization (e.g., DCT) and flattening, \proposed uses a unigram language model to calculate probabilities over the $n$-best token sequences for the chunk and then samples a token sequence. Tokenization sampling is performed only during training.}
    \label{fig:toast}
\end{figure*}

\textbf{Stochastic tokenization with frequency-based tokenizers.}
In natural language processing, subword tokenizers such as BPE~\cite{sennrich2016neural} and unigram language model~\cite{kudo2018subword} tokenization construct vocabularies from corpus statistics. Given a fixed vocabulary, the same underlying sequence can generally admit multiple valid segmentations, although standard tokenization selects a single one deterministically. Subword regularization~\cite{kudo2018subword} instead samples alternative tokenizations during training, and subsequent work has explored related stochastic tokenization strategies~\cite{provilko2020bpedropout, wang2021multiview,cognetta2024distributional,sims2026stochastok}. These methods have been shown to enable efficient learning in low-resource settings. Frequency-based action tokenizers such as FAST~\cite{pertsch2025fast} admit the same segmentation ambiguity, but whether sampling equivalent tokenizations improves robot policy learning has remained unexplored. This work investigates this question.

\section{Preliminaries}
\label{sec:preliminaries}
\textbf{Vision-language-action models.}
Given a language instruction $l$ and the current robot observation $o$, a VLA model predicts an $H$-step action chunk $a_{1:H}$. Each action $a_i \in \mathbb{R}^{D}$ consists of $D$ action dimensions, where $D$ depends on the robot embodiment and action representation. Autoregressive VLAs represent the action sequence as discrete tokens and model their distribution as $\pi_\theta(t_{1:N} \mid l,o)$. The predicted token sequence $t_{1:N}=t_1 \ldots t_{N}$ is decoded back into the continuous action sequence $a_{1:H}$ for execution. The action horizon $H$ is fixed by the policy configuration, whereas the number of tokens $N$ depends on the tokenizer and the action sequence.

\textbf{Action sequence tokenization and compression.}
Action sequence tokenization maps a continuous action chunk $a_{1:H}$ into a one-dimensional sequence of discrete tokens $t_{1:N}$. The action chunk is first transformed and discretized into $q \in \mathbb{Z}^{D \times H}$, for example by scalar binning or by applying DCT followed by discretization. The DCT represents each dimension's $H$-step trajectory by $H$ frequency coefficients, most of which become zero after scaling and rounding, making the flattened sequence highly compressible. The resulting discrete values are then flattened into a one-dimensional sequence. Timestep-major flattening orders the values as $q_{1,1} \ldots q_{D,1} q_{1,2} \ldots q_{D,H}$, whereas dimension-major flattening orders them as $q_{1,1} \ldots q_{1,H} q_{2,1} \ldots q_{D,H}$. With scalar binning~\cite{kim2024openvla}, each flattened discrete value directly corresponds to one action token, yielding $N=DH$. FAST~\cite{pertsch2025fast} instead applies byte pair encoding (BPE)~\cite{sennrich2016neural} to a DCT-transformed and discretized action sequence, producing a variable-length token sequence with $N \leq DH$.

\section{Method}
\label{sec:method}
\proposed constructs a unigram language model over quantized action sequences (\secref{subsec:consturcting_action_sequence_tokenizer}), samples alternative tokenizations that preserve the underlying action (\secref{subsec:stochastic_tokenization_of_action_chunks}), and uses the sampled token sequences for policy training (\secref{subsec:policy_training_with_toast}). The following sections describe each component in turn. \figref{fig:toast} shows the tokenization pipeline.

\subsection{Constructing Action Sequence Tokenizer}
\label{subsec:consturcting_action_sequence_tokenizer}
To enable stochastic tokenization of action chunks, \proposed first constructs a unigram language model over quantized action sequences. We use dimension-major flattening, which places temporally adjacent values of the same action dimension consecutively in the flattened sequence. Given a corpus of quantized and flattened action sequences $\mathcal{Q}=\{q^{1},\ldots,q^{M}\}$, defined in \secref{sec:preliminaries}, we estimate a vocabulary $\mathcal{V}$ whose tokens correspond to either individual quantized action values or contiguous subsequences of values. Following subword regularization~\cite{kudo2018subword}, we first construct a large seed vocabulary from candidate subsequences observed in $\mathcal{Q}$, while retaining all individual quantized values to ensure that any action sequence can be represented. The unigram model assigns each token $v \in \mathcal{V}$ a probability $p(v)$. The vocabulary and token probabilities are then iteratively optimized using the standard unigram language-model procedure, and low-contribution tokens are pruned until the target vocabulary size is reached. Given the resulting unigram language model, a flattened action sequence $q$ can generally be tokenized into multiple valid token sequences $t$. The probability of a token sequence is computed as $P(t) = \prod_{i=1}^{N} p(t_i)$, where $t_i$ denotes the $i$-th token in $t$. The unigram language model thereby assigns probabilities to alternative tokenizations of the same action sequence.

\begin{figure*}
    \centering
    \includegraphics[width=0.925\linewidth]{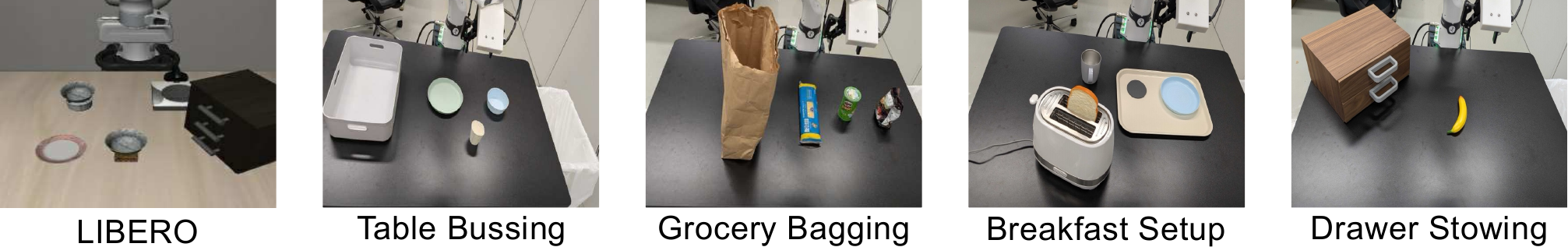}
    \caption{\textbf{Evaluation tasks.} We evaluate \proposed on the LIBERO benchmark and four real-robot manipulation tasks.}
    \label{fig:evaluation_tasks}
\end{figure*}

\begin{figure}[htbp]
  \centering
  \begin{minipage}[t]{0.49\linewidth}
    \centering
    \includegraphics[width=\linewidth]{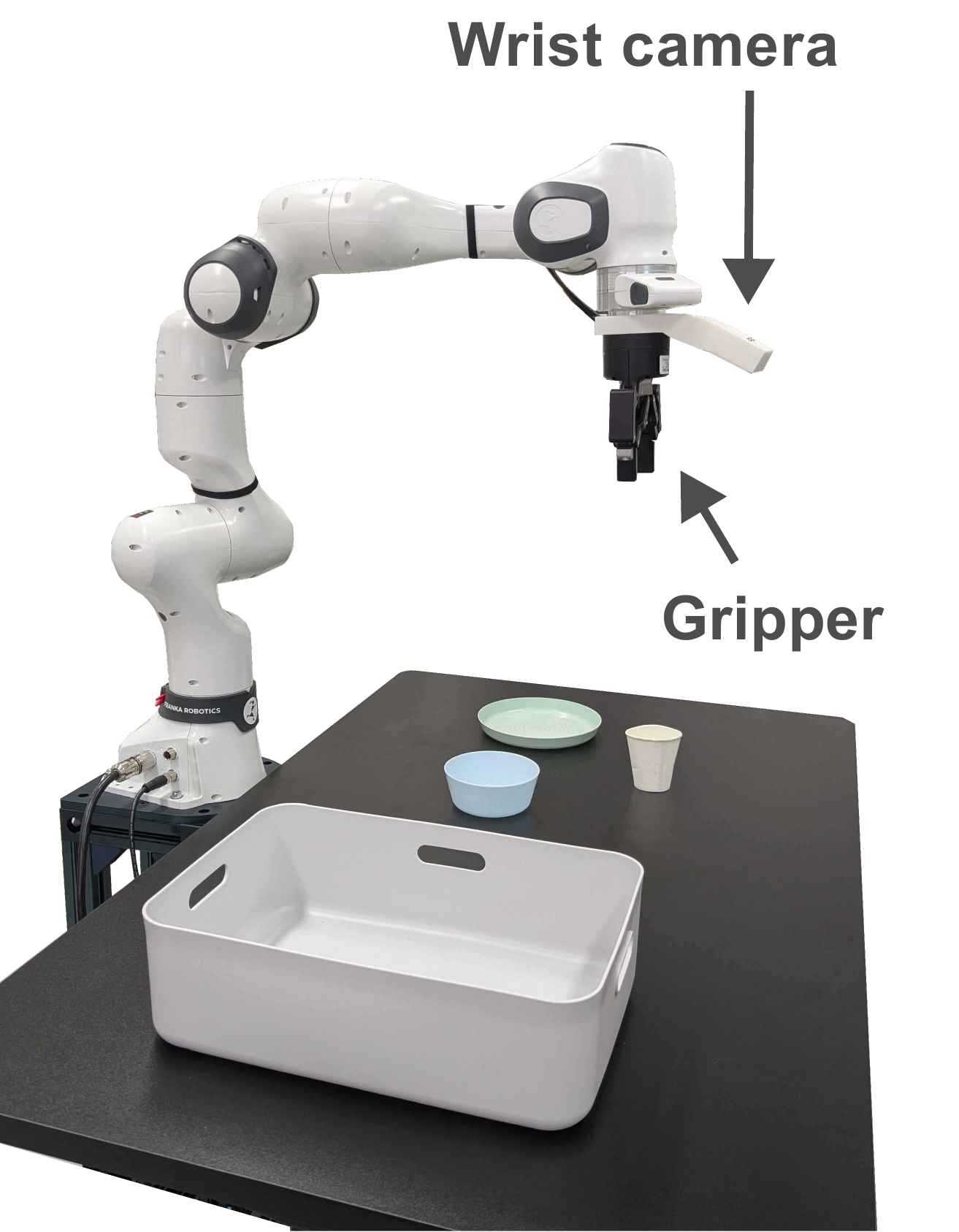}
    {\small (a) Wrist camera and Gripper.}
    \label{fig:wrist_camera}
  \end{minipage}
  \hfill
  \begin{minipage}[t]{0.49\linewidth}
    \centering
    \includegraphics[width=\linewidth, trim=0 35 0 0, clip]{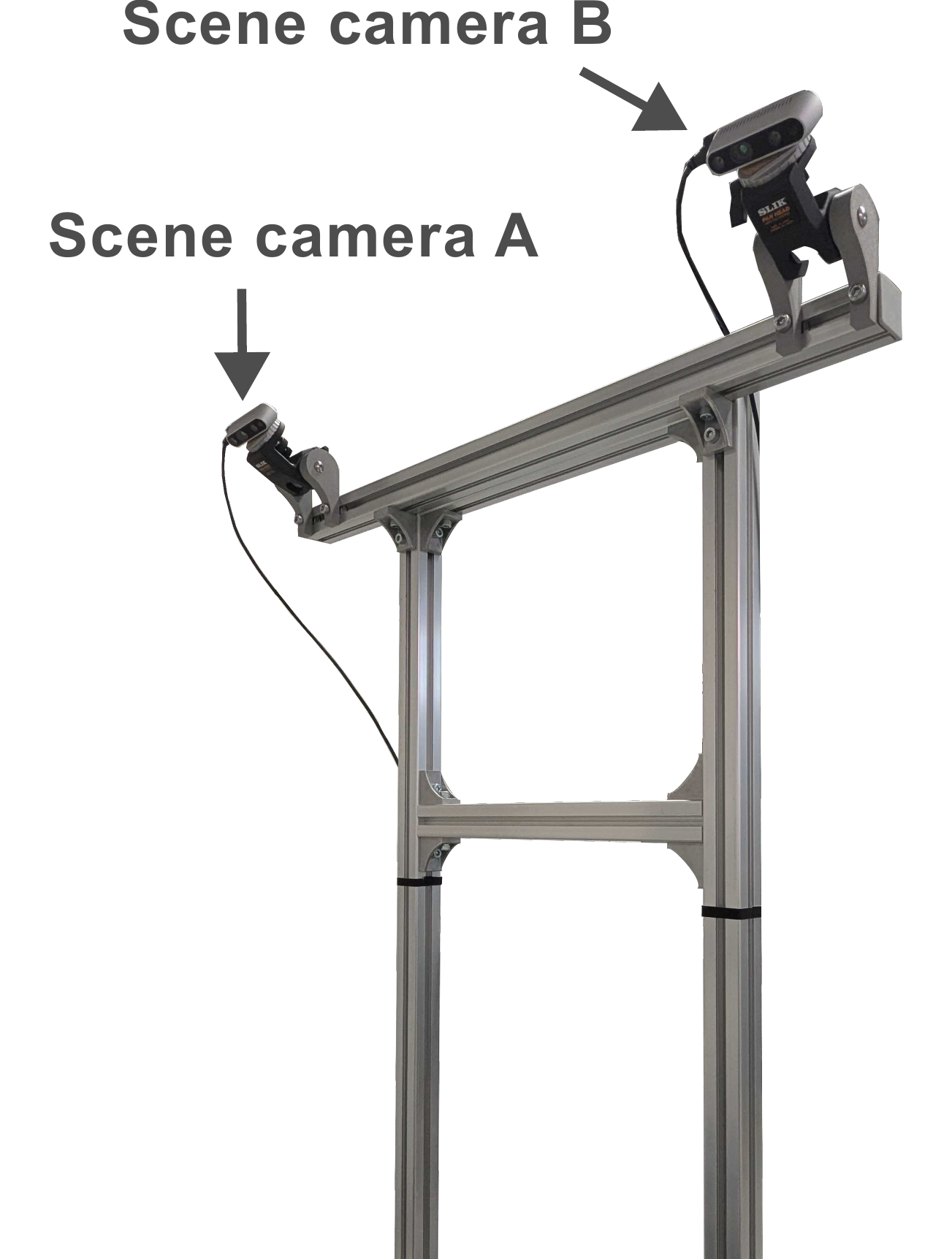}
    {\small (b) Two scene cameras.}
    \label{fig:third_person_cameras}
  \end{minipage}
  \\
  \vspace{1.5mm}
  \begin{minipage}{\linewidth}
    \centering
    \includegraphics[width=\linewidth]{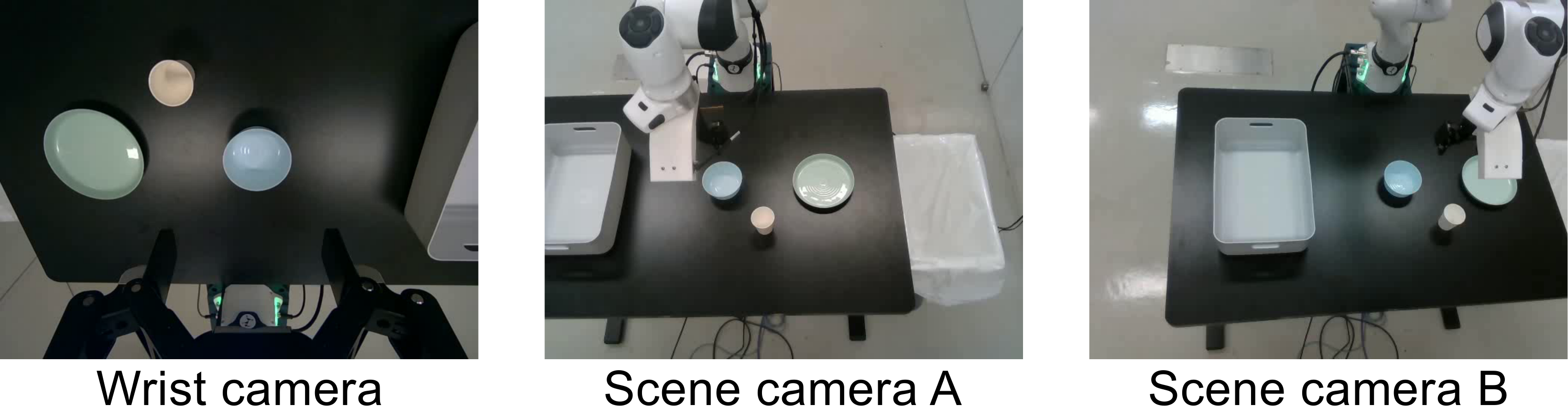}
    {\small (c) Camera views.}
    \label{fig:camera_views}
  \end{minipage}
  \caption{\textbf{Our real robot setup.} The robot (Franka Research 3) is equipped with a wrist camera (RealSense D405) and a Robotiq 2F-85 Gripper. Two scene cameras (RealSense D435) are placed facing the robot.}
  \label{fig:real_robot_setup}
  
\end{figure}

\subsection{Stochastic Tokenization of Action Chunks}
\label{subsec:stochastic_tokenization_of_action_chunks}
The unigram language model constructed in \secref{subsec:consturcting_action_sequence_tokenizer} allows a quantized and flattened action sequence ($q$) to be represented by multiple token sequences with different segmentation boundaries. While deterministic tokenization selects a single token sequence from these possible representations, \proposed probabilistically samples one of them. Given $q$, we obtain $n$ candidate token sequences $\{ t^1, t^2, \ldots, t^n \}$ by $n$-best search via unigram language model scores. Following subword regularization~\cite{kudo2018subword}, $n$-best search is performed in linear time by Forward-DP Backward-A* algorithm~\cite{nagata1994stochastic}, and the probability of the $i$-th candidate tokenization is approximated as
\begin{equation}
P(t^i \mid q) \simeq \frac{ P(t^i)^{\alpha} }{ \sum_{j=1}^{n} P(t^j)^{\alpha} },
\end{equation}
where $\alpha$ controls the sharpness of the multinomial distribution. A larger $\alpha$ leads to deterministic tokenization, while a smaller $\alpha$ leads to uniform sampling. Importantly, all sampled token sequences decode to the same quantized action sequence $q$. Thus, \proposed diversifies its discrete representation while preserving the underlying robot action. 

\subsection{Policy Training with \proposed}
\label{subsec:policy_training_with_toast}
In policy training with \proposed, a token sequence $\tilde{t}=\tilde{t}_1 \tilde{t}_2 \ldots \tilde{t}_{\tilde{N}}$ is sampled and used as the target at each training step. The policy is optimized to minimize the following standard next-token prediction objective:
\begin{equation}
-\sum_{i=1}^{\tilde{N}} \log \pi ( \tilde{t}_i \mid l,o,\tilde{t}_{<i} ).
\end{equation}
Thus, \proposed leaves the policy architecture and learning objective unchanged and modifies only the discrete representation used as the training target. 
At inference time, tokenization sampling is not applied. The policy autoregressively predicts an action token sequence, which is decoded into the corresponding continuous action sequence in the same manner as conventional token-based autoregressive policies.

\section{Experiments}
\label{sec:experiments}
We evaluate \proposed by addressing the following research questions: 
(\textbf{Q1}) Does stochastic tokenization improve policy learning, particularly when training data is limited?
(\textbf{Q2}) Does stochastic tokenization remain effective in real-world robot manipulation?
(\textbf{Q3}) How sensitive is \proposed to tokenizer design choices?
The remainder of this section explains experimental setup in \secref{subsec:experimental_setup} and answers \textbf{Q1}, \textbf{Q2}, and \textbf{Q3} in \secref{subsec:simulation_evaluation_across_data_scales}, \secref{subsec:real_robot_evaluation}, and \secref{subsec:analysis_of_tokenization_design_choices}, respectively.

\subsection{Experimental Setup}
\label{subsec:experimental_setup}
\textbf{Evaluation tasks.}
Evaluation is performed on both simulation and real-robot tasks (\figref{fig:evaluation_tasks}). We chose the \textbf{LIBERO} benchmark~\cite{liu2023libero} for simulation experiments. The benchmark features Franka Emika Panda with the control frequency of $10$ Hz. Actions are 7-dimensional values of a 6-DoF end-effector pose and an absolute gripper pose. LIBERO consists of four task suites: LIBERO-Spatial, LIBERO-Object, LIBERO-Goal, and LIBERO-Long~\cite{liu2023libero}.

The real-robot experiments feature Franka Research 3 with the control frequency of $20$ Hz, and we designed the following four manipulation tasks:
\begin{itemize}
    \item \textbf{Table Bussing} sorts trash on the desk into the trash bin and dishes into the dish rack, requiring accurate grasping of objects. 
    \item \textbf{Grocery Bagging} puts grocery items into a paper bag, requiring careful insertion of objects.
    \item \textbf{Breakfast Setup} moves a slice of toast onto a plate and a cup onto a coaster, requiring precise placement of objects at the intended locations.
    \item \textbf{Drawer Stowing} opens a drawer, puts an object inside, and closes it, requiring precise interaction with an articulated object.
\end{itemize}

The actions are represented as 8-dimensional values of 7-DoF joint angles and an absolute gripper pose. As shown in \figref{fig:real_robot_setup}, the robot was equipped with a single RealSense D405 as a wrist camera, two RealSense D435 as scene cameras, and a Robotiq 2F-85 gripper. 

\tabref{tab:amount_of_demos} shows the number of demonstrations collected for each task. We collected demonstrations by teleoperation with GELLO~\cite{wu2024gello}: 200 for Table Bussing, 200 for Grocery Bagging, 100 for Breakfast Setup, and 60 for Drawer Stowing. The number of demonstrations scales with the number of objects that the robot picks up in a task. \tabref{tab:average_episode_durations} shows average episode durations of the evaluation tasks. 

\begin{table}[t]
    \centering
    \caption{Amount of real-robot task demonstrations used for training policies.}
    \label{tab:amount_of_demos}
    \begin{tabular}{cccc}
        \hline
        Table   & Grocery & Breakfast & Drawer  \\
        Bussing & Bagging & Setup     & Stowing \\\hline
        200 & 200 & 100 & 60 \\\hline
    \end{tabular}
\end{table}

\begin{table}[t]
    \centering
    \caption{Average episode duration of the datasets (in seconds).}
    \begin{tabular}{cccccc}
        \hline
        \multirow{2}{*}{LIBERO} & Table & Grocery & Breakfast & Drawer \\
        & Bussing & Bagging & Setup & Stowing \\\hline 
        16.15 & 38.47 & 41.00 & 41.37 & 39.48 \\\hline
    \end{tabular}
    \label{tab:average_episode_durations}
\end{table}

\textbf{Policy configuration.}
All policies use PaliGemma-3B~\cite{beyer2024paligemma} as the backbone MLLM and are fine-tuned from vision-language-pretrained weights without prior action-policy training. This avoids biasing the comparison toward a particular action tokenizer through action-pretrained initialization. We implement all policies within the same autoregressive VLA pipeline based on OpenPI\footnote{\url{https://github.com/Physical-Intelligence/openpi}}. Actions are represented relative to the current proprioceptive state~\cite{chi2024umi}, except for the gripper state, which is represented in absolute coordinates. For each data scale, all methods are trained on identical demonstration subsets and use the same policy architecture and optimization settings. All policies are trained for 30,000 steps with a batch size of 32 using AdamW optimizer~\cite{loshchilov2018decoupled}. The action chunk horizon is set to $10$ for LIBERO and $20$ for the real-robot tasks to predict $1$-second action chunks. Action values are normalized to [-1, 1], with the bounds set to the 1st and 99th percentiles of each action dimension within each dataset. All results are produced in our pipeline rather than taken from prior papers.

\textbf{Tokenizer configuration.}
We compare \proposed with Binning~\cite{kim2024openvla}, FAST+ and FAST (rebuilt)~\cite{pertsch2025fast}, VQ-VLA~\cite{wang2025vqvla}, and BEAST~\cite{zhou2025beast}. We use VQ-VLA as the representative learned tokenizer because it is a residual VQ tokenizer with a public reference implementation. Unless otherwise specified, \proposed uses DCT quantization with a scaling factor of $10$, dimension-major flattening, and a unigram vocabulary of $512$ action tokens. The unigram tokenizer is implemented with SentencePiece~\cite{kudo2018sentencepiece}. For stochastic tokenization, we use $n=64$ candidate token sequences with $\alpha=0.1$ on LIBERO and $\alpha=0.5$ on the real robot.\footnote{$\alpha=0.5$ for the real robot was chosen in preliminary trials. As shown in \secref{subsec:analysis_of_tokenization_design_choices}, both values lie on the plateau, we do not expect this difference to affect the conclusions.} \proposed (deterministic) uses the same quantization, flattening, vocabulary, and policy configuration, but deterministically selects a single tokenization, thereby isolating the effect of stochastic tokenization.

Following FAST~\cite{pertsch2025fast}, which builds its vocabulary on a large external corpus, we construct the \proposed vocabulary from 1M DROID~\cite{khazatsky2024droid} action chunks (end-effector actions for LIBERO and joint angles for the real robot). An external corpus keeps the tokenizer fixed across data scales, so that only the policy training data varies in \secref{subsec:simulation_evaluation_across_data_scales}. Vocabularies built from the target demonstrations are examined in \secref{subsec:analysis_of_tokenization_design_choices}. FAST+ is the publicly released FAST tokenizer\footnote{\url{https://huggingface.co/physical-intelligence/fast}} and uses DCT quantization and BPE compression, with $2048$-token vocabulary built on 1M action chunks across multiple embodiments~\cite{pertsch2025fast}. For fair comparison, we also use FAST (rebuilt), which uses $512$-token vocabulary rebuilt from the same DROID corpus as \proposed. Binning discretizes each normalized action value in $[-1,1]$ into $256$ uniform bins, yielding $70$ tokens per $10\times7$ action chunk, and is fixed and data-independent. VQ-VLA uses the reference residual vector-quantized tokenizer with four $256$-code quantizers over non-overlapping $5$-step windows, yielding $8$ tokens per action chunk. Because a DROID-only VQ-VLA tokenizer produced poor LIBERO reconstruction and $0$\% downstream success rate, we train its codebook on DROID together with the corresponding LIBERO training data and refit it at each data scale to avoid using held-out trajectories. BEAST follows its LIBERO configuration with degree-$4$ B-splines, five basis functions, and $256$-bin quantization, yielding $35$ tokens per chunk. It requires no learned vocabulary and is therefore fixed across data scales. We use a fixed quantization range of $[-2,2]$ instead of the batch-statistics-based range in the reference implementation. 

\textbf{Evaluation protocol.}
On LIBERO, we follow the evaluation protocol of prior work~\cite{kim2024openvla}, evaluating $500$ rollouts per suite ($50$ per task) for each random seed, totaling $2,000$ rollouts. For the real-robot experiments, we evaluate $30$ rollouts per task for each random seed, with $5$ rollouts for each of $6$ task configurations varying object poses, colors, or locations. The same evaluation configurations are used across all methods. We report the mean success rate across three independently trained policies with different random seeds in both settings.
Significance is assessed with a hierarchical bootstrap that resamples seeds as clusters and episodes within each seed (10,000 resamples), pairing methods on identical evaluation episodes. A difference is marked significant when its two-sided $95$\% percentile CI excludes zero.

\subsection{Simulation Evaluation across Data Scales}
\label{subsec:simulation_evaluation_across_data_scales}
We first aim to address \textbf{Q1} through controlled experiments on LIBERO. To examine the performance of the policy under different amounts of training data, we train policies using $1/k$ ($k \in \{1,2,4,8,16\}$) of the available LIBERO demonstrations. For each data scale, subsets are drawn uniformly at random per suite with a fixed seed, and identical subsets are used across all methods.  The evaluation protocol is kept fixed regardless of the data scale.

\tabref{tab:simulation_results} shows the results. 
\proposed improves over its deterministic counterpart at every data scale, and the gain widens as training data is reduced, and it is significant with half or less of the data. Among the other deterministic baselines, the policies trained with \fast (rebuilt) and BEAST achieve performance comparable to that of the policy trained with \proposed when sufficient training data are available. As the amount of training data decreases, however, the performance gap between these baselines and \proposed becomes larger. With one quarter or less of the demonstrations, \proposed achieves the highest performance among the evaluated methods. Since \fast (rebuilt) uses the same quantization, tokenizer-construction data, and vocabulary size as \proposed, these results suggest that the improvement in the low-data settings is not due only to these factors, but also to stochastic tokenization. These results answer \textbf{Q1} affirmatively: stochastic tokenization improves policy learning, and its benefit grows as demonstrations become scarcer.

\begin{table}[t]
  \centering
  \caption{\textbf{LIBERO Results with different amounts of training data.} Mean success rate is calculated based on 2,000 rollouts per seed. * denotes that \proposed is significantly better than this row (the 95\% bootstrap CI of the difference excludes zero). This shows that stochastic action tokenization is more effective as available training data is reduced.}
  \label{tab:simulation_results}
  \begin{tabular}{lccccc}
  \hline
  \multicolumn{1}{l}{\multirow{2}{*}{Tokenizer}} & \multicolumn{5}{c}{Fraction of LIBERO training data} \\ 
  \cmidrule(lr){2-6}
  & $1/1$ & $1/2$ & $1/4$ & $1/8$ & $1/16$ \\\hline
  Binning~\cite{kim2024openvla} & 
    89.5\phantom{*} &
    80.0* & 
    70.7* & 
    56.4* & 
    37.8* \\ 
  FAST+~\cite{pertsch2025fast} & 
    85.7* & 
    59.4* & 
    40.6* & 
    30.1* & 
    17.8* \\ 
  FAST (rebuilt)~\cite{pertsch2025fast} & 
    \textbf{92.4}\phantom{*} & 
    83.8\phantom{*} & 
    71.3* & 
    55.7* & 
    34.4* \\ 
  BEAST~\cite{zhou2025beast} & 
    91.5\phantom{*} & 
    82.6\phantom{*} & 
    69.8* & 
    55.2* & 
    37.1* \\ 
  VQ-VLA~\cite{wang2025vqvla} & 
    82.8* & 
    74.7* & 
    58.3* & 
    46.8* & 
    31.6* \\ 
  \cmidrule(lr){1-6}
  \proposed (deterministic) & 
    90.9\phantom{*} & 
    82.7* & 
    69.8* & 
    55.0* & 
    36.0* \\ 
  \proposed & 
    91.4\phantom{*} & 
    \textbf{84.9}\phantom{*} & 
    \textbf{75.3}\phantom{*} & 
    \textbf{60.2}\phantom{*} & 
    \textbf{42.8}\phantom{*} \\ 
  \hline
  \end{tabular}
\end{table}

\begin{figure*}[t]
    \centering
    \includegraphics[width=\linewidth]{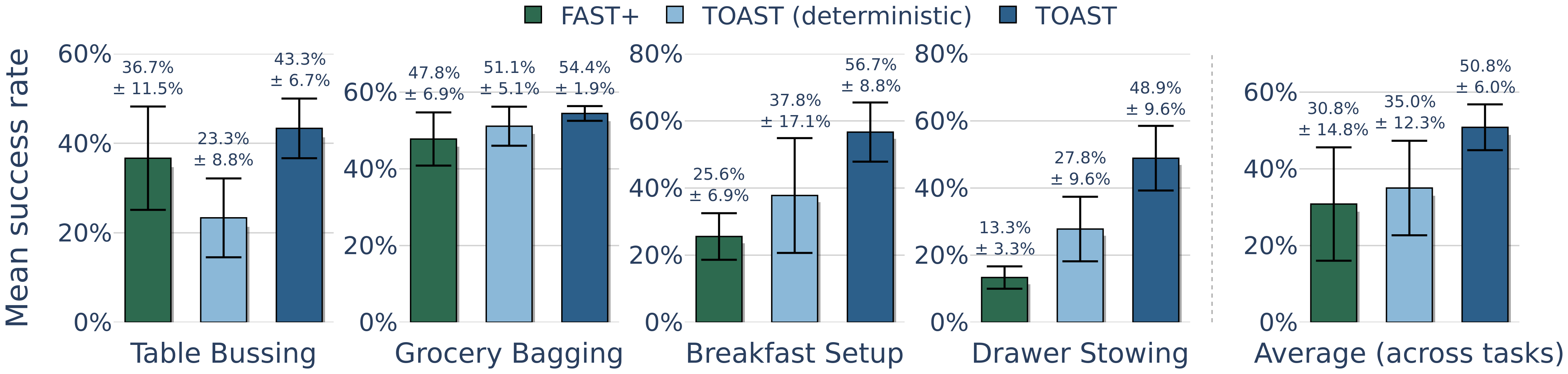}
    \caption{\textbf{Real-robot task results.} We evaluate policies with three different tokenizations: \proposed, \proposed (deterministic), and FAST+. Each policy is evaluated based on $30$ rollouts per task with three random seeds. \proposed achieves the highest success rate (50.8\%), improving performance of \proposed (deterministic) by 15.8 points and FAST+ by 20.0 points.}
    \label{fig:real_robot_results}
\end{figure*}

\subsection{Real-Robot Evaluation}
\label{subsec:real_robot_evaluation}
We next address \textbf{Q2} through real-robot experiments. The primary comparison is between \proposed and its deterministic counterpart, which isolates the effect of stochastic action tokenization under identical configurations. We additionally include FAST+ as an external reference because it is a publicly released deterministic tokenizer based on the same frequency-based compression paradigm as \proposed. FAST (rebuilt) differs from TOAST (deterministic) only in BPE vs. unigram segmentation and the flattening direction, and the two are within 1.6 points at every data scale in \tabref{tab:simulation_results}. We therefore regard TOAST (deterministic) as a proxy for FAST (rebuilt) on the real robot. The policy decodes tokens greedily and maps them back to a $D \times H$ chunk by vocabulary lookup and inverse DCT; on the real robot, generating a 1-second (20-step) joint-angle chunk takes 168 ms on a single RTX 4090. Chunks are executed open-loop: the robot tracks the 20 Hz targets with a 1 kHz joint impedance controller using zero-order hold, and keeps executing the current chunk while the next one is generated. If a predicted sequence cannot be decoded into a complete chunk, it is discarded and inference is re-run on the next observation. This rarely occurs, and the same pipeline is applied identically to all tokenizers.

\figref{fig:real_robot_results} shows the results on the real-robot tasks. We observe that \proposed outperforms \proposed (deterministic) by $15.8$ points and FAST+ by $20.0$ points, and both differences are significant under the same bootstrap protocol as in \secref{subsec:simulation_evaluation_across_data_scales}. While stochastic tokenization yields larger improvements in the real-robot experiments than in the full-data simulation setting, this result is consistent with the data-scaling trend observed on LIBERO, where the benefit of sampling increases as the amount of training data decreases. Since real-robot demonstrations are substantially more limited than those in simulation, these results further support the practical value of stochastic action tokenization in data-constrained real-world scenarios. Since the simulation and real-robot evaluations also differ in task distribution, action representation, and execution conditions, we do not attribute the larger real-robot gains to data scale alone.

The results also show that larger gains from stochastic tokenization are obtained in all tasks except Grocery Bagging. During the rollouts, we observed that the deterministic baselines frequently fail on actions that require higher precision, such as grasping the rim of a plate or accurately reaching a drawer handle. These errors often lead directly to task failure, whereas the policy trained with stochastic tokenization executes such interactions more reliably. This observation suggests that the larger real-robot gains are associated in part with improved robustness in action segments that require precise manipulation.

\subsection{Analysis of Tokenization Design Choices}
\label{subsec:analysis_of_tokenization_design_choices}
We next address \textbf{Q3} by analyzing the design and behavior of stochastic action tokenization. Specifically, we examine the sensitivity to the sampling parameter $\alpha$, the corpus used to construct the vocabulary, the flattening order, and the policy's sensitivity to equivalent tokenizations, and finally report the training overhead of sampling.

\begin{figure}[t]
    \centering
    \includegraphics[width=\linewidth]{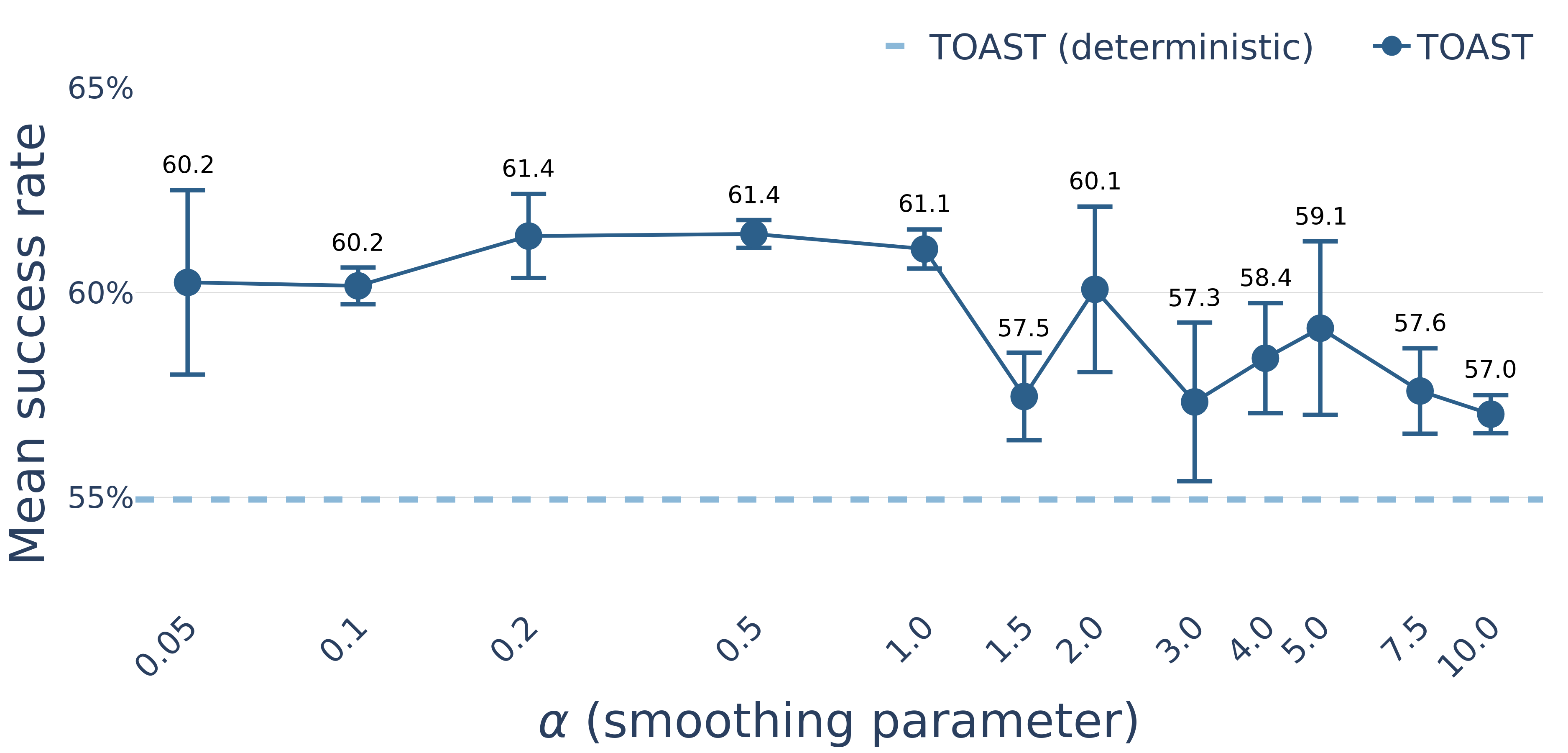}
    \caption{\textbf{LIBERO ($1/8$) results with varying sampling hyperparameter $\alpha$.} Performance is stable for $\alpha \in [0.05, 1.0]$ and degrades toward the deterministic baseline for larger $\alpha$.}
    \label{fig:alpha_sweep}
\end{figure}

\textbf{Sensitivity to the sampling hyperparameter $\alpha$.}
We first examine the sensitivity of stochastic tokenization to the smoothing parameter $\alpha$, which controls the concentration of the sampling distribution over alternative tokenizations. We conduct this analysis on LIBERO using $1/8$ of the training demonstrations, where the effect of stochastic tokenization is pronounced. As shown in \figref{fig:alpha_sweep}, policy performance remains stable over a broad range of $\alpha$ from $0.05$ to $1.0$ and consistently exceeds the deterministic counterpart. Increasing $\alpha$ concentrates the sampling distribution toward high-probability tokenizations, approaching deterministic tokenization, as in subword regularization~\cite{kudo2018subword}. Consistent with this behavior, policy performance generally decreases at larger $\alpha$, although the trend is not strictly monotonic.

\textbf{Effect of the vocabulary corpus.}
Our default tokenizer vocabulary is built from DROID, an external corpus that is much larger than the policy training data. To examine whether the benefit of stochastic tokenization depends on this external corpus, we construct vocabularies from DROID+LIBERO and from LIBERO only, where the LIBERO portion is restricted to the same $1/k$ subset used for policy training, and keep all other settings unchanged. \tabref{tab:simulation_results_across_vocabs} shows the results. With both vocabularies, \proposed significantly outperforms its deterministic counterpart at every data scale, including full data. Notably, the deterministic counterpart barely changes across vocabularies (e.g., $36.0$, $36.4$, and $37.9$ at $1/16$ for DROID, DROID+LIBERO, and LIBERO only), whereas the gain from sampling grows as the vocabulary is built closer to the target domain ($+6.8$, $+9.6$, and $+11.9$, respectively). We attribute this to the vocabulary containing more tokens that match the target action patterns, which yields more plausible alternative segmentations for sampling to exploit. We keep the DROID vocabulary as the default to keep the tokenizer independent of the policy training data and consistent with the FAST (rebuilt) comparison. The results here indicate that this is a conservative choice. Thus, the low-data benefit of stochastic tokenization does not rely on an external corpus and persists, and in fact grows, even when the vocabulary is constructed solely from the target demonstrations.

\begin{table}[t]
\centering
\caption{\textbf{Effect of the vocabulary corpus.} LIBERO success rate of \proposed and \proposed (deterministic) with vocabularies built from DROID+LIBERO and only LIBERO. The LIBERO portion is restricted to the same $1/k$ subset used for policy training. $^*$: \proposed is significantly better than \proposed (deterministic).}
\label{tab:simulation_results_across_vocabs}
\begin{tabular}{@{}lccccc@{}}
\toprule
\multirow{2}{*}{Tokenizer} & \multicolumn{5}{c}{Fraction of LIBERO training data} \\
\cmidrule(l){2-6}
 & 1/1 & 1/2 & 1/4 & 1/8 & 1/16 \\
\midrule
\multicolumn{6}{@{}l}{\textit{(Vocabulary: DROID + LIBERO)}} \\
\proposed (deterministic) & 
    90.1* & 
    81.8* & 
    69.2* & 
    53.0* & 
    36.4* \\
\proposed & 
    92.2\phantom{*} &
    85.3\phantom{*} &
    74.7\phantom{*} &
    61.3\phantom{*} &
    46.0\phantom{*} \\
\midrule
\multicolumn{6}{@{}l}{\textit{(Vocabulary: LIBERO)}} \\
\proposed (deterministic) & 
    89.1* & 
    80.3* & 
    67.5* & 
    53.8* & 
    37.9* \\
\proposed & 
    91.3\phantom{*} &
    83.8\phantom{*} &
    73.2\phantom{*} &
    65.0\phantom{*} &
    49.8\phantom{*} \\
\bottomrule
\end{tabular}
\end{table}

\textbf{Effect of the flattening direction.}
Our default dimension-major flattening places temporally adjacent values of the same action dimension consecutively, whereas timestep-major flattening in previous work~\cite{kim2024openvla,pertsch2025fast} arranges all action dimensions at each timestep consecutively. We examine the impact of stochastic tokenization on different flattening directions using LIBERO. Concretely, we compare performance improvements by sampling (success rate gain of \proposed from \proposed (deterministic)) across different amounts of LIBERO training data in both flattening directions.

\begin{figure}[t]
    \centering
    \includegraphics[width=\linewidth]{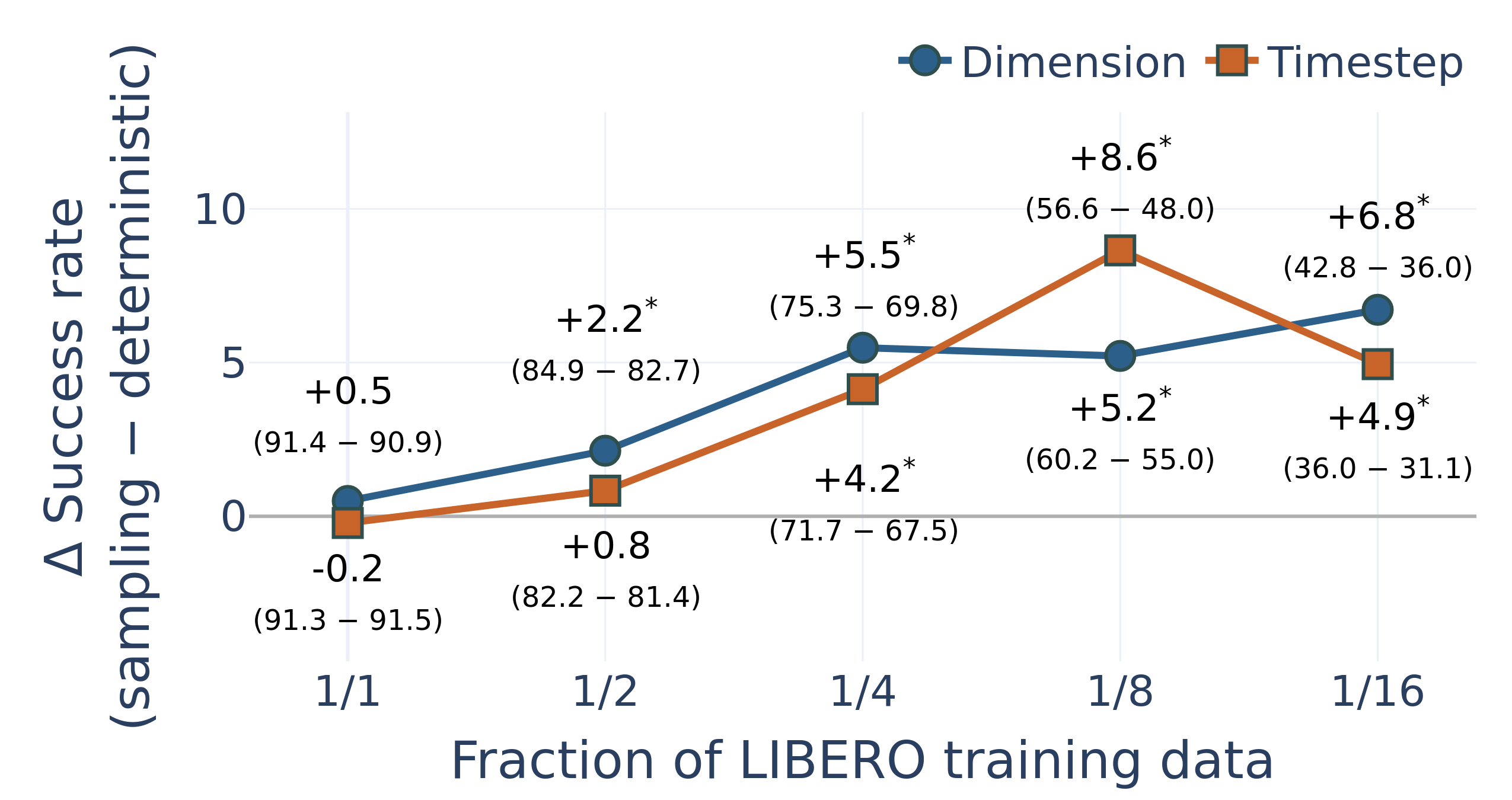}
    \caption{\textbf{Success rate gain by sampling.} * denotes a significant improvement over the deterministic baseline. In both flattening directions, stochastic action tokenization tends to achieve more improvements when limited data is available.}
    \label{fig:sampling_gain_by_data_sizes}
\end{figure}

\figref{fig:sampling_gain_by_data_sizes} shows success rate gains by sampling in both flattening directions. Dimension-major flattening tends to achieve higher performance improvement than timestep-major flattening in our experiments, and it achieves $6.8$ points higher absolute performance at $1/16$ scale ($42.8\%$ by dimension vs $36.0\%$ by timestep), supporting its use as the default configuration, while the two flattenings are on par (91.4 vs. 91.3) at full data. The larger gain of timestep-major flattening at $1/8$ (+$8.6$) reflects its lower deterministic baseline ($48.0$ vs. $55.0$) rather than a higher absolute performance ($56.6$ vs. $60.2$). More importantly, however, the benefit of stochastic tokenization is not specific to this flattening choice. Sampling improves performance with timestep-major flattening, with clear gains when fewer training data are available. Thus, although the flattening order affects absolute performance, sampling improves both representations with gains that grow as data shrink (significant at 1/4 and below in both directions), indicating that the low-data benefit observed in \secref{subsec:simulation_evaluation_across_data_scales} is not an artifact of the flattening order used by \proposed.

\begin{figure}[t]
    \centering
    \includegraphics[width=\linewidth]{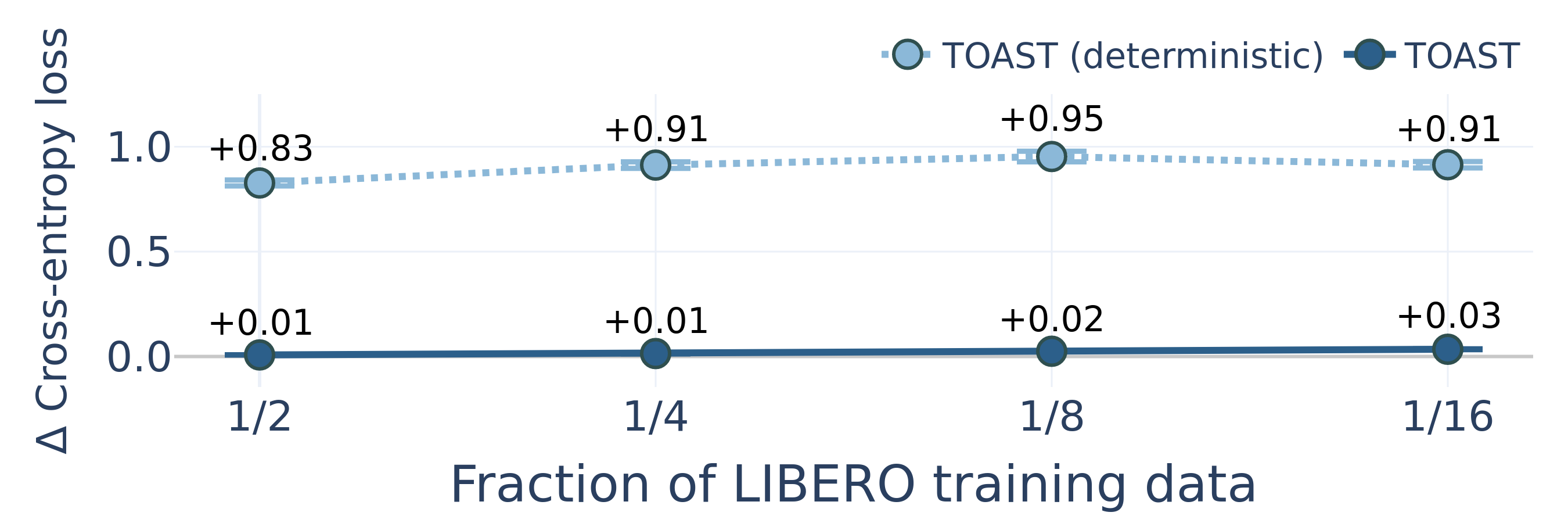}
    \caption{
    \textbf{Sensitivity to equivalent tokenizations ($\Delta_{\mathrm{tok}} = L_{\mathrm{alt}} - L_{\mathrm{can}}$) on 100 held-out LIBERO demonstrations.} Stochastic tokenization reduces dependence on a single tokenization of an action chunk.}
    \label{fig:tokenization_sensitivity}
\end{figure}

\textbf{Sensitivity to equivalent action tokenizations.}
We examine whether stochastic tokenization reduces the policy's dependence on a particular tokenization of an action sequence. A frequency-based tokenizer can represent the same quantized action sequence using multiple valid token sequences, although deterministic training exposes the policy to only one of them. For each held-out LIBERO action chunk, we compute the cross-entropy loss for its canonical tokenization and for $63$ alternative tokenizations sampled from the same unigram model. Since all of these tokenizations decode to the same quantized action sequence, their loss difference measures sensitivity to the choice of discrete representation. We define this quantity as $\Delta_{\mathrm{tok}}=L_{\mathrm{alt}}-L_{\mathrm{can}}$, where $L_{\mathrm{alt}}$ is averaged over the alternative tokenizations. We use this metric as a representation-level diagnostic, rather than as a proxy for closed-loop policy performance.

As shown in \figref{fig:tokenization_sensitivity}, deterministic training produces a consistently large positive loss gap across data scales, indicating a strong preference for the canonical tokenization used during training. In contrast, stochastic training reduces this gap to nearly zero, showing less sensitivity to how the same underlying action is segmented into tokens. This is not because stochastic training inflates the loss overall. On held-out data, \proposed attains lower $L_{\mathrm{can}}$ than its deterministic counterpart (1.146 vs. 1.252 at $1/8$), indicating better generalization.
We observe the same qualitative behavior on the training trajectories, suggesting that this difference reflects the representation learned by the policy rather than a held-out generalization effect. These results confirm that stochastic training has the intended effect of reducing dependence on a single tokenization of an action sequence.

\textbf{Computational cost.} 
Averaged over $1,000$ training steps on each of the five tasks, one step takes 0.417 s with \proposed, 0.417 s with \proposed (deterministic), and 0.415 s with FAST+ (0.4\% increase). This demonstrates that $n$-best search runs in linear time and its sampling cost is negligible.

\section{Limitations}
\label{sec:limitations}
First, \proposed relies on multiple valid segmentations of the same quantized sequence and therefore does not directly apply to learned vector-quantized tokenizers~\cite{wang2025vqvla,liu2026faster,liu2026oat}. Extending stochastic tokenization to learned discrete representations is left for future work. Second, our experiments use a single Franka embodiment with end-effector actions in simulation and joint-angle actions on the real robot. We do not study a shared vocabulary across embodiments as in \cite{pertsch2025fast,liu2026faster}. Third, we instantiate stochastic tokenization with a unigram model and subword regularization~\cite{kudo2018subword}. Other stochastic tokenization schemes such as BPE-dropout~\cite{provilko2020bpedropout} were not compared, and whether the benefit stems from the unigram model specifically or from stochastic segmentation in general remains open. Finally, the vocabulary size (512) and the number of candidates $n$ were fixed throughout, and their sensitivity is not examined.

\section{Conclusion} 
\label{sec:conclusion}
We presented \proposed, a stochastic action sequence tokenization method for autoregressive robot policies. By sampling multiple tokenizations that represent the same quantized action sequence during training, \proposed provides diverse discrete supervision without additional robot demonstrations. Experiments on LIBERO show that stochastic tokenization consistently improves over its deterministic counterpart, with larger gains when fewer demonstrations are available. We observe consistent improvements across diverse real-robot manipulation tasks. Analyses show that these gains are robust to the flattening order, sampling parameter, and vocabulary corpus, while stochastic training substantially reduces the policy's sensitivity to a particular tokenization of the same action sequence. These results support stochastic action tokenization as an effective approach for learning autoregressive robot policies from limited demonstrations.

\section{Acknowledgment}
This work was supported by the JST CREST project Real-World Intelligent Systems, ``MORAL: Morphoception-Oriented Reasoning and Action with Language'' (JPMJCR2553) and FRONTia, a Japanese national program led by the Ministry of Economy, Trade and Industry of Japan (METI) and the New Energy and Industrial Technology Development Organization (NEDO), aimed at developing a domestic multimodal foundation model for AI robots and physical AI. 
This work is also supported by JSPS KAKENHI Grant Number 25K21274. We thank Koshi Makihara for their support with the experimental setup.

{
    \bibliographystyle{IEEEtran}
    \bibliography{root}
}


\appendix

\subsection{Hyperparameters for Constructing \proposed Tokenizer}
\label{app:hyperparameters}
\tabref{tab:hyperparameters_tokenizer} summarizes the hyperparameters of \proposed tokenizer construction used in \secref{subsec:simulation_evaluation_across_data_scales} and \secref{subsec:real_robot_evaluation}. In both LIBERO and real robot setups, the tokenizers adopt dimension-major flattening of quantized actions. The tokenizer construction corpus consists of 1M action chunks sampled from DROID, and end-effector poses and joint angles are used for LIBERO and DROID, respectively, to match the action representation the policy predicts.

\subsection{Hyperparameters for Policy Training}
\tabref{tab:hyperparameters_tokenizer} shows the hyperparameters for training policies used in \secref{subsec:simulation_evaluation_across_data_scales} and \secref{subsec:real_robot_evaluation}. The policy parameters are tuned using AdamW~\cite{loshchilov2018decoupled}. Actions in continuous action chunks are represented as relative values to the current proprioceptive states~\cite{chi2024umi}. In order to be robust to outliers, action values are normalized within the range $[-1, 1]$ via quantile normalization, following prior work~\cite{pertsch2025fast}.

\subsection{Task Evaluation Rubric}
\label{app:task_evaluation_rubric}
In real-robot evaluation in \secref{subsec:real_robot_evaluation}, we use a task evaluation rubric and judge if the rollout is successful or not. The following lists a rubric for each task.

\begin{itemize}
\item \textbf{Table Bussing} requires the robot to sort three objects on a table according to their type. Tableware items (e.g., plates and bowls) must be placed in the dish rack, while disposable items (e.g., paper cups) must be placed in the trash can. A rollout is considered successful if all objects are placed in their designated locations within 120 seconds; otherwise, it is considered a failure. 

\item \textbf{Grocery Bagging} requires the robot to place three objects into a paper bag. The manipulation targets are a bag of pasta, a bag of coffee, and a tube of chips. A rollout is considered successful if all objects are placed into the paper bag within 120 seconds without knocking it over; otherwise, it is considered a failure. 

\item \textbf{Breakfast Setup} requires the robot to place a coffee cup onto a coaster and transfer a slice of toast from a toaster onto a plate. A rollout is considered successful if both objects are placed in their designated locations within 120 seconds; otherwise, it is considered a failure. Additionally, if the robot lifts the toaster together with the toast, we stop the robot for safety reasons and mark the rollout as a failure.

\item \textbf{Drawer Stowing} requires the robot to open a drawer, move an object from the table into the drawer, and then close the drawer. The manipulation target is a bowl, a banana, or a plum. A rollout is considered successful if, within 120 seconds, the object is inside the drawer and the  drawer is fully closed; otherwise, it is considered a failure.
\end{itemize}

\begin{table}[t]
    \centering
    \caption{Hyperparameters for \proposed.}
    \begin{tabular}{lcc}
        \hline
        Hyperparameter & LIEBRO & Real Robot (FR3) \\
        \hline
        Vocabulary size & 512 & 512 \\
        $\alpha$ (smoothing parameter) & 0.1 & 0.5 \\
        $n$ ($n$-best search) & 64 & 64 \\
        \hline
    \end{tabular}
    \label{tab:hyperparameters_tokenizer}
\end{table}

\begin{table}[t]
    \centering
    \caption{Hyperparameters used for training policies.}
    \begin{tabular}{lcc}
        \hline
        Hyperparameter & LIBERO & Real Robot (FR3) \\
        \hline
        Training steps & 30,000 & 30,000 \\
        Warmup steps & 1,000 & 1,000 \\
        Batch size & 32 & 32 \\
        Learning rate & $2.5 \times 10^{-5}$ & $2.5 \times 10^{-5}$ \\
        Horizon & 10 & 20 \\
        \hline
    \end{tabular}
    \label{tab:hyperparameters_training}
\end{table}

\subsection{Successful Rollout Visualization on Real-Robot Tasks}
\begin{figure*}[htbp]
  \centering
  \begin{subfigure}{\textwidth}
    \centering
    \includegraphics[width=0.9\linewidth]{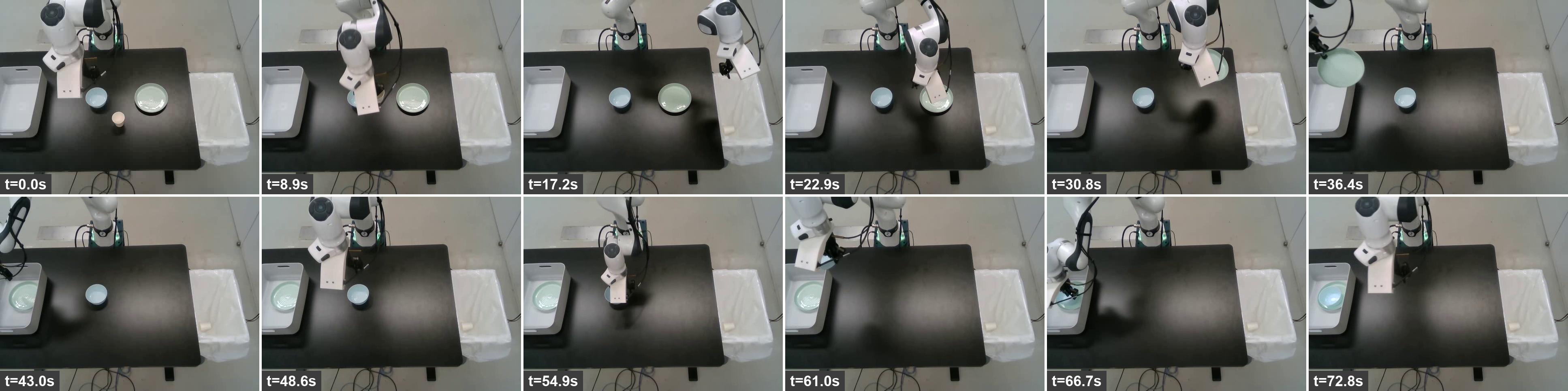}
    \caption{Table Busing.}
  \end{subfigure}
  
  \begin{subfigure}{\textwidth}
    \centering
    \includegraphics[width=0.9\linewidth]{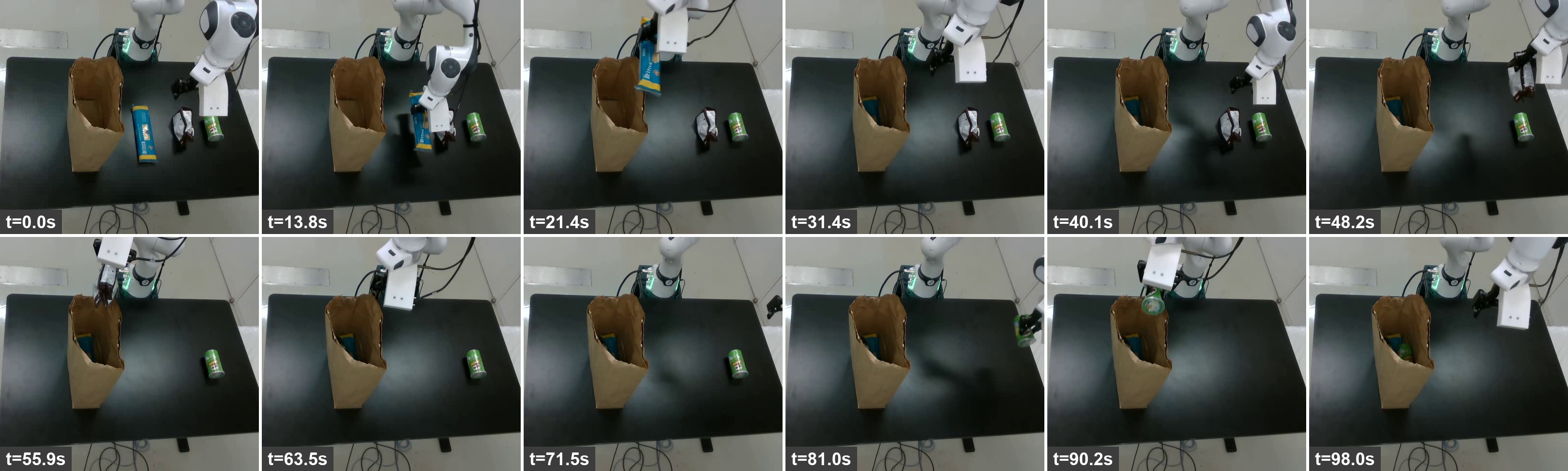}
    \caption{Grocery Bagging.}
  \end{subfigure}
  
  \begin{subfigure}{\textwidth}
    \centering
    \includegraphics[width=0.9\linewidth]{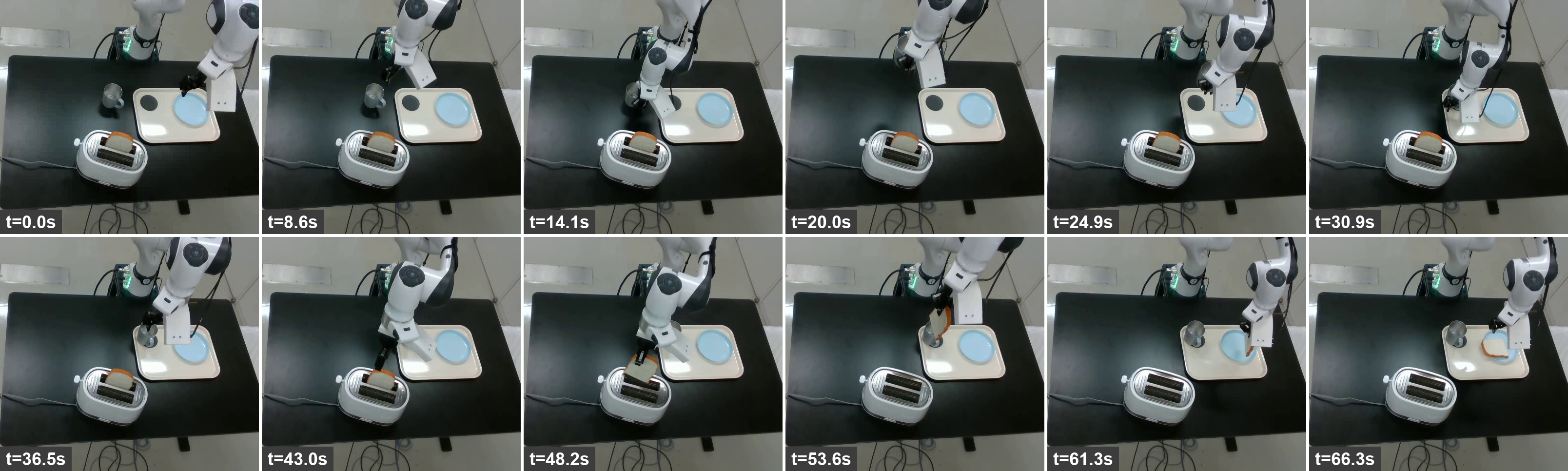}
    \caption{Breakfast Setup.}
  \end{subfigure}
  
  \begin{subfigure}{\textwidth}
    \centering
    \includegraphics[width=0.9\linewidth]{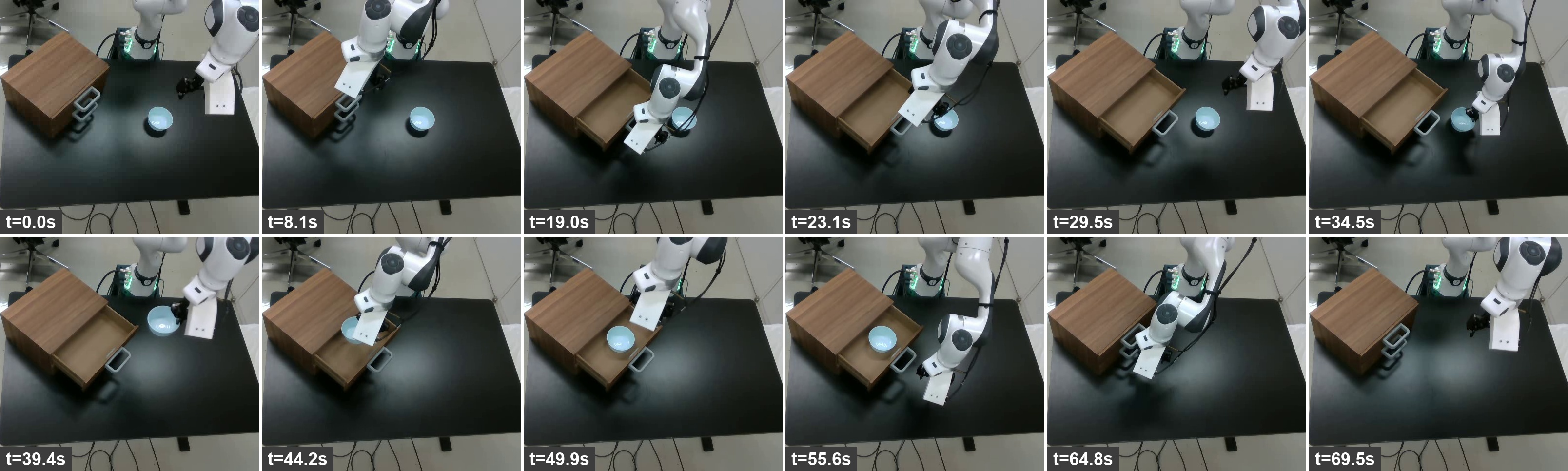}
    \caption{Drawer Stowing.}
  \end{subfigure}
  
  \caption{Successful rollouts on real-robot tasks.}
  \label{fig:real_robot_task_visualization_success}
\end{figure*}

\figref{fig:real_robot_task_visualization_success} shows the keyframes from successful rollouts of policies trained with the proposed \proposed tokenizer. The first frame ($t=0.0$s) in each task corresponds to the initial state of the environment.

\end{document}